\documentclass[11pt]{article}
\usepackage{verbatim}
\usepackage{graphicx}
\usepackage{amsmath}
\usepackage[square,numbers]{natbib}
\usepackage[nonatbib, preprint]{neurips_2023}
\usepackage[utf8]{inputenc} 
\usepackage[T1]{fontenc}    
\usepackage{hyperref}       
\usepackage{url}            
\usepackage{booktabs}       
\usepackage{amsfonts}       
\usepackage{nicefrac}       
\usepackage{microtype}      
\usepackage{wrapfig}
\usepackage{color}
\usepackage[list=true]{subcaption}

\title{CapText: Large Language Model-based Caption Generation From Image Context and Description}

\author{
  Shinjini Ghosh$^*$ \\
  Computer Science and \\
  Artificial Intelligence Laboratory \\
  Massachusetts Institute of Technology \\
  \And 
  Sagnik Anupam$^*$ \\
  Computer Science and \\
  Artificial Intelligence Laboratory \\
  Massachusetts Institute of Technology \\
\\}

\begin{document}
\maketitle
\def\thefootnote{*}\footnotetext{Equal contribution}

\begin{abstract}
While deep-learning models have been shown to perform well on image-to-text datasets, it is difficult to use them in practice for captioning images. This is because \textit{captions} traditionally tend to be context-dependent and offer complementary information about an image, while models tend to produce \textit{descriptions} that describe the visual features of the image. Prior research in caption generation has explored the use of models that generate captions when provided with the images alongside their respective descriptions or contexts. We propose and evaluate a new approach, which leverages existing large language models to generate captions from textual descriptions and context alone, without ever processing the image directly. We demonstrate that after fine-tuning, our approach outperforms current state-of-the-art image-text alignment models like OSCAR-VinVL on this task on the CIDEr metric.
\end{abstract}

\section{Introduction}

\begin{wrapfigure}{R}{0.5\textwidth}
    \centering
    \includegraphics[width=0.48\textwidth]{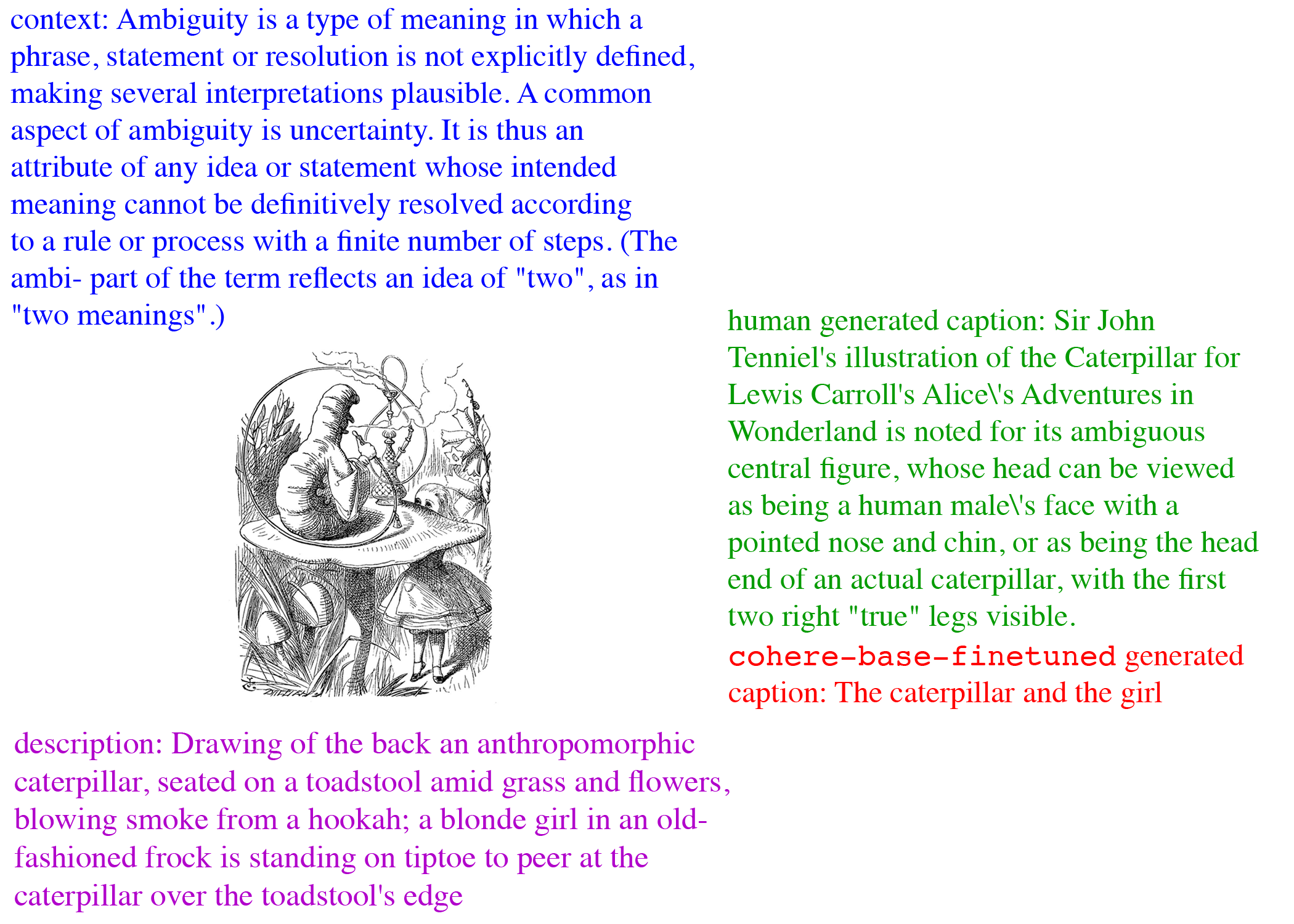}
    \caption{Caption generation for an illustration from Alice in Wonderland, taken from \url{https://commons.wikimedia.org/wiki/File:Alice_05a-1116x1492.jpg}}
    \label{fig:alice}
\end{wrapfigure}

Deep learning models have been shown to have excellent performance on image annotation tasks as well as several benchmark image-to-text datasets. However, there is a crucial limitation that has held back the adoption of language models for image captioning purposes - not enough emphasis has been placed in those datasets on the communicative purpose of the caption \citep{kreiss2021concadia}. Thus, to help develop image-captioning models for more practical use cases, researchers have begun to distinguish between image \textit{descriptions}, which are texts that are intended to replace the images in their current context, and image \textit{captions}, which are texts that are intended to accompany the images alongside their current context. An example of an image's description, context, human-generated caption, and a model-generated caption is provided in Figure \ref{fig:alice}.

While image-description generation has been shown to be a much easier task than caption generation, it has been observed that including representations of the textual context alongside the image representation improves performance on both description and caption-generation tasks. These results have been obtained with two kinds of models. Firstly, hybrid models comprising a computer vision model and a language model (like ResNet-LSTM and DenseNet-LSTM) have been used on this task, where the ResNet/DenseNet's representation of the image is passed to the LSTM alongside the BERT embeddings of the context. Secondly, the current state-of-the-art performance on this task has been shown by the Transformer-based vision-language model OSCAR, where the VinVL pre-trained visual feature extractor provides a concatenation of the encodings of visual features, object tags, and context as input to the transformer model \citep{kreiss2021concadia}.

In this paper, we present a novel approach that leverages the power of large language models and their understanding of context to examine if it might be possible for a large language model to generate a good caption without ever seeing an image (or its encoded representation) in the first place. Our approach provides large language models with the description of the image alongside the paragraph next to which the image appears (which we shall refer to as the \textit{context} of the image). We then instruct our models to generate context-appropriate captions for the image. We hypothesize that this approach will lead to higher-quality captions due to two reasons. Firstly, large-language models do not need to worry about extracting high-quality features from images since we are providing the large-language models only with textual data, and thus there is no noise from the image encodings. Secondly, since large-language models are trained on large corpora of text extracted from the web, we believe that they can leverage their learned representations to provide additional context that will improve the quality of the captions. We then validate our hypothesis by comparing our model-generated captions to the existing state-of-the-art models using CIDer scores.

We evaluate our approach using three models: Cohere's \texttt{base} model (hereby referred to as \texttt{cohere-base}, formerly known as \texttt{xlarge}) and Open AI's \texttt{text-davinci-003} (GPT 3.5), as well as the open-source GPT-2. We first evaluate zero-shot performance on our two models by simply providing the description and the context of the image in a single prompt and asking the model to generate captions. Both models perform better than the DenseNet-LSTM and Resnet-LSTM image-text alignment models, but worse than the Oscar-VinVL state-of-the-art CIDEr score of 1.14.

We then evaluate if finetuning on a small training dataset leads to any difference in performance, and fine-tune \texttt{cohere-base} on a small dataset of 100 examples. We discover that the fine-tuned \texttt{cohere-base} model achieves a CIDEr score of 1.729, beating state-of-the-art performance by models processing images. We believe that our approaches would allow the captioning technique to be deployed at scale with relative ease since there is no need to process the actual images themselves, and thus would speed up the deployment of the models.

The rest of the paper is structured in the following manner: we first describe existing approaches to image captioning in Section \ref{sec:2}. Then, in Section \ref{sec:3}, we provide an overview of the Concadia dataset. We then describe our prompts and finetuning approaches in Section \ref{sec:4}, where we also provide an overview of the CIDEr metric and its suitability for the captioning task. We describe some limitations of our approach, alongside a description of the implications of deploying our models in Section \ref{sec:5}, and present our conclusions and some direction for future research in Section \ref{sec:6}.

\section{Related Works}
\label{sec:2}

\subsection{Concadia: Towards Image-Based Text Generation with a Purpose}

\subsubsection{Introduction}
This paper pertains to the primary dataset we aim to utilize for evaluating our approaches \citep{kreiss2021concadia}. The authors claim that current state-of-the-art deep learning models often achieve great results on the benchmark image-to-text datasets but fail to generate practically useful texts. They postulate that this gap can be bridged by distinguishing \textit{descriptions} from \textit{captions} because they hold separate communicative roles. The \textit{Concadia} dataset consists of 96,918 images with their corresponding English descriptions, captions, and surrounding context. The dataset is also relatively new; thus, we believe this would be the perfect one to utilize for our tasks.

\subsubsection{Problem Addressed}

Despite their ability to generate coherent and fluent text in various contexts, image-based natural language generation techniques have limitations in their practical usefulness for tasks such as accessibility, image search, creative writing, and navigational instructions, e.g., \citep{dognin2019adversarial, Guinness2018, gurari2020captioning}. While they show promise for these applications, they currently do not produce outputs that are sufficiently useful for these tasks. This paper focuses on identifying and tackling a major obstacle that hinders the performance of these systems. Specifically, the inadequate consideration of the communicative intention of the generated text is highlighted and addressed in the context of the distinction between \textit{descriptions} and \textit{captions}. The former focuses on the visual features and is intended to replace an image, oftentimes due to accessibility, while the latter focuses on complementing the information already provided by an image by supplying additional information. The authors also provide evidence from a pre-registered human-subjects experiment which further reinforces the basic assumption about the distinct communicative purposes of descriptions and captions.

\subsubsection{Solution Proposed}

The authors introduce the Concadia dataset, a Wikipedia-based corpus of 96,918 images with associated English-language captions, alternate text descriptions, and surrounding context from the respective Wikipedia article. Apart from reporting a thorough linguistic analysis of the similarities and differences between the descriptions and captions of images in Wikipedia, the authors also develop and investigate image-based natural language generation systems using Transformer-based and LSTM-based architectures. They observe that given the closer descriptive relationship between images and their corresponding descriptions, it is easier for models to generate descriptions rather than captions. This result also inspired us to take on the challenge of caption generation in our own research. We also note that the exact counterpart of this result can be seen in prompts to Dall-E \cite{ramesh2021zeroshot}, where prompting a description yields a much closer image to what one desires, rather than a caption-like prompt. The other finding in the \cite{kreiss2021concadia} paper was that it is highly beneficial to include representations of the textual context in which the image appeared. We also took this finding to heart, thus including the surrounding text in our experiments as well.

\subsubsection{Impact and Significance}

This provides researchers with an opportunity to directly draw comparisons and contrasts between descriptions and captions for images. One could also employ it to build image-based natural language generation systems that fulfill these specific communicative objectives. In our work, we utilize both the unique position of this dataset in the field as well as the well-curated information it provides, to introduce a new approach for caption generation from surrounding text and descriptions, without ever directly involving the image itself.

\subsection{Is a Caption Worth a Thousand Images?}\label{sec:concadia}

\subsubsection{Introduction}
This 2022 paper on ``A Controlled Study for Representation Learning" provides a background for why our proposed work might yield promising results \cite{santurkar2022caption}. The authors postulate that the only irreducible difference between CLIP \cite{radford2021learning} and SimCLR \citep{chen2020simple} is whether supervision is provided to the model via image augmentations or image-caption matching {---} and they aim to study the same. This comparison reveals that neither form of supervision (using images alone or coupled with language) is strictly better than the other. They also discovered some practical applications where CLIP's performance was unable to be matched using SimCLR with \textit{any} amount of image data and also other cases where the language supervision was harmful. Overall, this paper provides a very nice discussion of a question very adjacent to our problem statement.

\subsubsection{Problem Addressed}
Image-based contrastive learning approaches have been promising in building generalized models by leveraging large-scale unlabelled data sources. However, \citep{radford2021learning} showed in their 2021 paper how contrastive learning with language supervision can yield CLIP models which possess pretty remarkable transfer capabilities. This sparked a debate on the utility of multi-modality in representation learning among the vision and natural language processing communities, of which this paper seeks to address the specific question of whether language supervision leads to more transferable representations than using just images. The authors further state that while, at first glance, the answer may seem obvious, given that CLIP utilized caption information unavailable to traditional image-based approaches as well as showed significant improvement over prior work in \cite{radford2021learning}. Nonetheless, CLIP stands out from these methods in numerous aspects, including the training data and the specific implementation choices that are made at a detailed level \cite{devillers-etal-2021-language}. These differences pose a challenge in discerning the sole impact of language guidance on the performance of the model.

\subsubsection{Solution Proposed}
The authors developed a methodology for evaluating the effectiveness of language supervision in representation learning. They recognized that both CLIP and popular image-based methods use contrastive learning as their underlying primitive. They found that given that CLIP and SimCLR \cite{chen2020simple} have many similarities in their conceptual framework, the main difference between them is the way they are provided with supervision. While SimCLR is supervised through image augmentations, CLIP is supervised through image-caption matching. The authors systematically compared matched versions of SimCLR and CLIP to evaluate the value of language supervision in downstream transfer performance. They observe that the results primarily depended on three properties of the pre-training dataset. Large-scale datasets with image-caption pairs result in better CLIP performance compared to SimCLR. However, in low-data regimes, language supervision can negatively impact model performance. The descriptiveness \cite{kreiss2021concadia} of dataset captions determines how well the resulting CLIP models transfer, and a single descriptive image-caption pair (e.g., from \cite{Lin2014}) is worth more than multiple less descriptive captions (e.g., from YFCC \cite{Thomee_2016}). Additionally, caption variability within a dataset can harm CLIP's performance. The authors propose a modification to CLIP training, which involves performing text data augmentations by sampling from a pool of captions for each image to address this issue. Based on their research findings, they propose straightforward interventions that can improve the transferability of CLIP models on datasets. Firstly, they suggest filtering out low-quality captions with the aid of a text-based classifier. Secondly, they recommend applying data augmentation to captions by using pre-trained language models to paraphrase them \cite{wang2021gptj6b}.

\subsubsection{Impact and Significance}

This work is significant in its contribution to the ongoing debate on the role of language in visual representation learning. The authors' comparison of CLIP with matched image-only SimCLR models reveals that neither form of supervision is strictly better than the other and that the utility of language supervision is dependent on various properties of the pre-training dataset. Through their analysis, the authors were able to identify algorithmic improvements and dataset modifications that better take advantage of language supervision, providing practical implications for future research and development in this area.

Despite some limitations in their exploration, including the scope of their analysis and some remaining inconsistencies, the authors' results suggest that they successfully isolated crucial confounders in the comparison of CLIP and SimCLR. Additionally, their work highlights the potential risks of incorporating biased or stereotyped language into visual representation learning, raising important ethical considerations for the development and use of such models. Overall, this work provides important insights into the role of language in visual representation learning and lays the foundation for further research in this area, including ours.

\subsection{VinVL: Revisiting Visual Representations in Vision-Language Models}

\subsubsection{Introduction}
This 2021 paper offers a new model on the vision-language side of the field \citep{zhang2021vinvl}. It is an improved object detection model to provide object-centric representations of images. They pre-train a large-scale object-attribute detection model based on the X152-C4 architecture. It achieves much better results than the then-state-of-the-art models on a wide range of VL tasks. They also present a comprehensive empirical study to demonstrate that the visual features matter in VL models as well as provide a detailed ablation study of their pre-trained object detection model. They also claim that their model can generate visual features for a richer collection of visual objects and concepts that are crucial for VL tasks, which serve as guiding principles in our own project.

\subsubsection{Problem Addressed}
Existing vision language (VL) research focuses heavily on improving the vision-language fusion model and leaves a lot to be desired as far as object detection model improvements are concerned. They also aimed to fill the gap left by the lack of a detailed ablation study of pre-trained object detection models.

\subsubsection{Solution Proposed}
The authors \textit{(i)} present a comprehensive empirical study to demonstrate that visual features matter in VL models, \textit{(ii)} develop a new object detection model that can produce better visual features of images as compared to the classical OD model \cite{anderson2018bottomup} and significantly improves state-of-the-art results on all major VL tasks across multiple public benchmarks, \textit{(iii)} provide a detailed ablation study of their pre-trained object detection model to investigate the relative contribution of the performance improvement due to the different design choices. 

\subsubsection{Impact and Significance}
Alongside generating a very impactful insight regarding the utility of visual features in VL models, the authors also provide a significantly improved state-of-the-art model which we have utilized as one of the benchmark models in our research as well.

\section{The Concadia Dataset}

\label{sec:3}

\begin{wrapfigure}{R}{0.5\textwidth}
    \centering
    \includegraphics[width=0.48\textwidth]{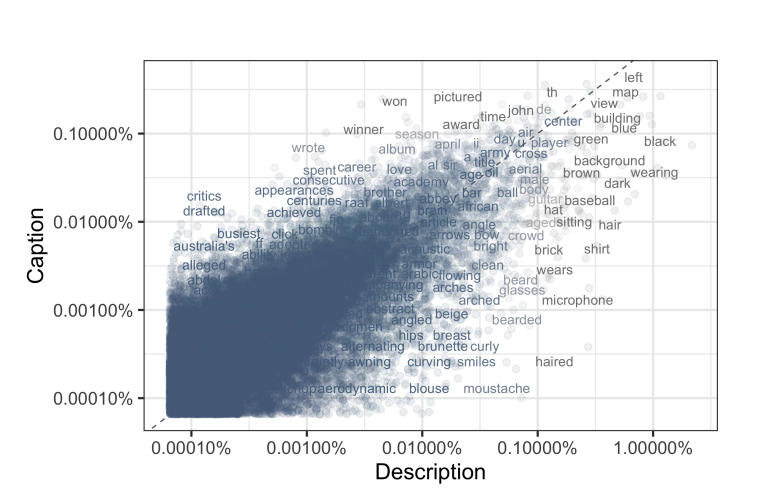}
    \caption{Correlation of word frequency between captions and descriptions in the Concadia dataset (Figure 10 from \cite{kreiss2021concadia})}
    \label{fig:concadiacorr}
\end{wrapfigure}

We primarily use the Concadia dataset \cite{kreiss2021concadia}, which we also detail in Section \ref{sec:concadia}, to evaluate our approaches. The Concadia dataset comprises 96,918 images from Wikipedia, and their associated context, descriptions, and captions. The associated context here is defined to be the surrounding text from the 41,143 articles from which these images are taken. A train/test/dev split is also provided by the authors. Images that appear multiple times with different associated texts, and images with identical captions and descriptions are specifically placed into the training set to ensure that the validation and test sets are of the highest quality. Any other assignments are randomized while maintaining the constraint that data from the same article are assigned to the same split. We evaluate our models directly on the test split of the original Concadia dataset for direct comparison with the performance of the other image-text alignment models. The test split has 9691 data points from 4951 articles, with a mean caption length of 12.67 words and description length of 14.29 words. Moreover, alongside a thorough analysis of bigrams and trigrams and their different occurrences across captions and descriptions, the authors also provide a direct comparison of which words occur more frequently in captions and descriptions, which is shown in Figure~\ref{fig:concadiacorr}.

\section{Methods}

\label{sec:4}

\subsection{Models}

We evaluate our models on the test split of the Concadia dataset, which comprises 9691 data points. For each data point, we combined the image descriptions (present in the alt text on the original Wikipedia pages) with the textual context surrounding the image (from the Wikipedia article). The prompt we used for caption generation was as follows: \texttt{"Here is a description of an image: \{\textcolor{red}{description}\} \textbackslash n Here is the context the image appears in: \{\textcolor{red}{context}\} \textbackslash n. An appropriate caption for the image is: "}.

We evaluated our approaches using three models: \texttt{cohere-base}, \texttt{text-davinci-003} (GPT-3.5), and GPT-2. \texttt{cohere-base} (formerly known as \texttt{xlarge}) is a closed-source model by the Canadian startup Cohere with a context window of 2048 tokens. As per the Stanford HELM website, \texttt{xlarge} has 52.4B parameters as of 2022-06-09. \texttt{text-davinci-003} is a 175B parameter closed-source model by Open AI, which leverages Proximal Policy Optimization (PPO) \cite{schulman2017proximal} in a manner similar to InstructGPT \cite{ouyang2022training} to use reward models trained from human comparisons. GPT-2 is a 1.5B parameter open-source Transformer model with a context window of 1024 tokens, first introduced by OpenAI in 2019 \cite{radford2019language}. 

We evaluate the relative performance of the approaches and models using the CIDer metric \citep{vedantam2015cider}, which has been shown to outperform both BLEU \citep{papineni2002bleu} and ROUGE \citep{lin2004rouge} metrics. CIDEr is designed to capture the semantic similarity between generated captions and reference captions more accurately, being a consensus-based approach that takes semantic diversity and the variability of reference captions into account, whereas BLEU and ROUGE are based on n-gram overlap between the generated and reference captions. Furthermore, CIDEr uses TF-IDF weighting to account for the relative importance of different words in the captions, which helps to address the problem of common n-grams dominating the evaluation and thus has been shown to have a stronger correlation with human judgments of caption quality compared to BLEU and ROUGE. Additionally, as per our dataset statistics, 26\% of Concadia's caption data and 15\% of its description data comprises proper names, making other metrics like SPICE \citep{anderson2016spice} and METEOR \citep{denkowski2014meteor} ill-suited choices, especially as is evidenced by the thorough experiments conducted in \cite{vedantam2015cider} comparing the CIDEr metric with the BLEU, ROUGE, and METEOR metrics.

\subsection{Evaluation Metric: CIDEr}

CIDEr \cite{vedantam2015cider} is a popular evaluation metric for image captioning tasks. It measures the similarity between a candidate caption and a set of reference captions using consensus-based statistics.

The CIDEr score is based on the idea of consensus: if many captions use similar words and phrases to describe an image, then those words and phrases are likely to be more relevant and accurate descriptions of an image $I_i$. Conversely, if few captions use a particular word or phrase, it is less likely to be an accurate or relevant description. The goal of this metric is to automatically evaluate how well a candidate sentence $c_i$ matches the consensus of a set of image descriptions $S_i = \{s_{i1}, \ldots, s_{im}\}$. The authors argue that intuitively, a measure of the aforementioned consensus would encode how often the $n$-grams ($\omega_k$s) in $c_i$ are present in the reference sentences.

To encode this intuition, they perform a Term Frequency Inverse Document Frequency (TF-IDF) weighting for each $n$-gram \cite{Robertson2004}. The number of times an $n$-gram $\omega_k$ occurs in a reference sentence $s_{ij}$ is denoted by $h_k(s_{ij})$ or $h_k(c_i)$ for the candidate sentence $c_i$. The TF-IDF weighting $g_k(s_{ij})$ is computed for each $n$-gram $\omega_k$ using the following equation, where $\Omega$ is the vocabulary of all $n$-grams and $I$ is the set of all images in the dataset. The first term measures the TF of each $n$-gram $\omega_k$, and the second term measures the rarity of $\omega_k$ using its IDF.

\begin{equation*}
\begin{gathered}
    g_k(s_{ij}) \\
    = \frac{h_k(s_{ij})}{\sum_{\omega_l\in\Omega}h_l(s_{ij})}
    \log\frac{|I|}{\sum_{I_p\in I}\min\left(1,\sum_q h_k(s_{p,q})\right)}
\end{gathered}
\end{equation*}

The CIDEr$_n$ score for $n$-grams of length $n$ is thus computed using the average cosine similarity between the candidate sentence and the reference sentences, thus accounting for both precision and recall. Here, where $g^n(ci)$ is a vector formed by $g_k(ci)$ corresponding to all $n$-grams of length $n$ and $||g^n(ci)||$ is the magnitude of the vector $g^n(ci)$, and similarly for $g^n(s_{ij})$.

\begin{align*}
& \text{CIDEr}_n(c_i, S_i) \\
& = \frac{1}{m} \sum_j \frac{g^n(c_i) \cdot g^n(s_{ij})}{\lVert g^n(c_i)\rVert\lVert g^n(s_{ij})\rVert}
\end{align*}

The authors use higher order, i.e., longer $n$-grams (of up to 4) to capture the grammatical properties as well as the richer semantics of the captions, and combine the scores from $n$-grams of varying length as follows. They find that uniform weights of $w_n = 1/N$ work the best and use $N=4$.

\begin{equation*}
\text{CIDEr}(c_i,S_i) = \sum_{n=1}^N w_n\text{CIDEr}_n(c_i,S_i)
\end{equation*}

We compute the CIDEr score over the large Concadia dataset of images and captions, evaluating our generated captions with the human-generated captions for each image. 

\section{Results and Discussion}
\label{sec:5}

For both models, we initially attempted to generate image captions in a zero-shot manner, by simply supplying the prompt to the base models without any subsequent fine-tuning on the downstream task. We discovered that while both models outperformed the ResNet-LSTM and DenseNet-LSTM baselines on the CIDEr metric, they were unable to beat the state-of-the-art performance of the OSCAR-VinVL model. Seeking to improve performance, we fine-tuned the better-performing of the two models (\texttt{cohere-base}) on 100 random data points from the training split of the original Concadia dataset. The CIDEr scores are presented in Table \ref{Tab:2}.

\begin{wraptable}{l}{0.5\textwidth}
\begin{tabular}{cc}
     \hline
     Model & CIDEr \\
     \hline
     \textbf{cohere-base-finetuned} & \textbf{1.73} \\
     Oscar-VinVL & 1.14 \\
     \textbf{text-davinci-003 (GPT-3.5)} & \textbf{1.12} \\
     \textbf{cohere-base} & \textbf{0.62} \\
     \textbf{GPT-2} & \textbf{0.51} \\
     DenseNet-LSTM & 0.19 \\
     ResNet-LSTM & 0.17 \\
     \hline
\end{tabular}
\caption{CIDEr scores comparing model-generated captions to human-generated captions from Concadia's test split for the different models. Models in \textbf{bold} only received the context and the description of the image as input, while the unbolded models received the image and the description of the image as input.}
\label{Tab:2}
\end{wraptable}

As evidenced by the CIDEr scores of the models, it is difficult for our models to outperform the existing state-of-the-art image captioning models without any finetuning. We know that the Oscar-VinVL model is trained on a dataset comprising of BERT text embeddings of the image's description and its object tags concatenated with the visual features of the image. Hence, we hypothesize that this difficulty in outperforming the existing state-of-the-art occurs because Oscar-VinVL is able to learn to generate high-quality captions from the text embeddings. It does not struggle with aligning text with the image's embeddings the way the LSTMs do with the other models, since the image-text alignment component of the text is outsourced to the VinVL model, which generates helpful object tags. However, after fine-tuning, the \texttt{cohere-base} model is able to generalize well to the problem of generating high-quality captions by leveraging its existing knowledge representations.

\section{Limitations}
\label{sec:6}

\begin{wrapfigure}{R}{0.5\textwidth}
    \centering
    \includegraphics[width=0.48 \textwidth]{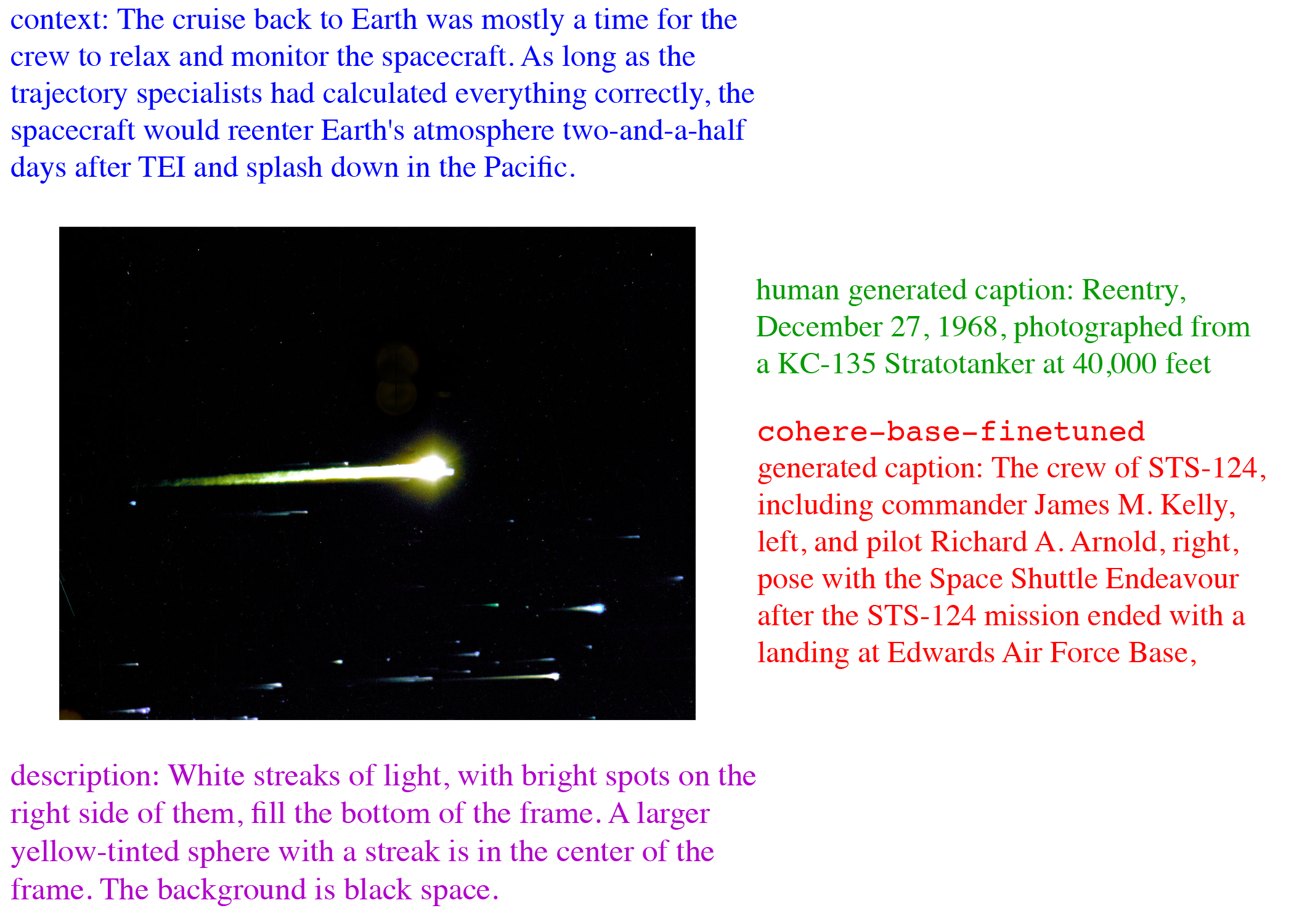}
    \caption{Caption generation for the Apollo 8 re-entry image, taken from \url{https://commons.wikimedia.org/wiki/File:Apollo_8_reentry,_December_27,_1968.jpg}}
    \label{fig:apollo}
\end{wrapfigure}

A serious limitation of our approach is the inability to fact-check the generated captions for truthfulness, due to the constraints of the dataset and the models used. For example, in Figure \ref{fig:apollo}, the fine-tuned Cohere model makes several errors. While it is able to guess the general subject domain of the photograph (and the corresponding article), it mislabels the Apollo 8 re-entry photograph (from 1968) as a photograph of astronauts standing next to the Space Shuttle Endeavor at the Edwards Air Force Base after the STS-124 mission in 2008. It then also goes on to name the individuals involved in the STS-124 mission, citing James M. Kelly as the mission commander and Richard A. Arnold as the pilot. 

However, the mission commander of the STS-124 mission was \textit{Mark E.} Kelly, while the pilot was \textit{Kenneth T. Ham}. An astronaut named Richard \textit{R.} Arnold was the mission specialist on the STS-119 mission in 2009. Both STS-124 and STS-119 were flown on the Space Shuttle Discovery, and landed at the Kennedy Space Center in Florida. Thus, in this instance, not only did the model caption confuse the photograph for a mission 40 years later, it generated incorrect information about the mission's crew, space shuttle, and landing site. For future research, it is crucial to fact-check the model's generations by incorporating modern fact-checking methods, such as attribution techniques like RARR \cite{gao2022attributed} or factuality-assessment techniques like SelfCheckGPT \cite{manakul2023selfcheckgpt}, into the model's text generation pipeline.

\section{Conclusions}
\label{sec:7}

By leveraging the generalization ability of large language models for caption generation, our approach of letting large language models only access the image's description and its context outperformed the existing state-of-the-art Oscar-VinVL model on the CIDEr metric after finetuning. We also presented the results of our attempts of leveraging zero-shot learning techniques to teach the model to generate high-quality captions. Finally, we discussed some of the limitations of our approaches, even if they perform better than the state-of-the-art, and suggested some harm mitigation techniques which can be used to improve the quality of the results.

In the future, we hope to examine the quality of captions generated by providing large language models with machine-generated image descriptions, such as VinVL's object tags, as opposed to the human-written alt-text present in the original dataset. If we are able to demonstrate that the quality of summaries does not go down significantly upon doing so, we will be able to build an end-to-end image captioning system that is able to generate useful captions simply by using the image and its surrounding textual context as input.

\bibliography{custom}

\begin{thebibliography}{}

\bibitem[Anderson et~al., 2016]{anderson2016spice}
Anderson, P., Fernando, B., Johnson, M., and Gould, S. (2016).
\newblock Spice: Semantic propositional image caption evaluation.
\newblock In {\em Computer Vision--ECCV 2016: 14th European Conference,
  Amsterdam, The Netherlands, October 11-14, 2016, Proceedings, Part V 14},
  pages 382--398. Springer.

\bibitem[Anderson et~al., 2018]{anderson2018bottomup}
Anderson, P., He, X., Buehler, C., Teney, D., Johnson, M., Gould, S., and
  Zhang, L. (2018).
\newblock Bottom-up and top-down attention for image captioning and visual
  question answering.

\bibitem[Chen et~al., 2020]{chen2020simple}
Chen, T., Kornblith, S., Norouzi, M., and Hinton, G. (2020).
\newblock A simple framework for contrastive learning of visual
  representations.
\newblock In {\em International conference on machine learning}, pages
  1597--1607. PMLR.

\bibitem[Denkowski and Lavie, 2014]{denkowski2014meteor}
Denkowski, M. and Lavie, A. (2014).
\newblock Meteor universal: Language specific translation evaluation for any
  target language.
\newblock In {\em Proceedings of the ninth workshop on statistical machine
  translation}, pages 376--380.

\bibitem[Devillers et~al., 2021]{devillers-etal-2021-language}
Devillers, B., Choksi, B., Bielawski, R., and VanRullen, R. (2021).
\newblock Does language help generalization in vision models?
\newblock In {\em Proceedings of the 25th Conference on Computational Natural
  Language Learning}, pages 171--182, Online. Association for Computational
  Linguistics.

\bibitem[Dognin et~al., 2019]{dognin2019adversarial}
Dognin, P.~L., Melnyk, I., Mroueh, Y., Ross, J., and Sercu, T. (2019).
\newblock Adversarial semantic alignment for improved image captions.

\bibitem[Gao et~al., 2022]{gao2022attributed}
Gao, L., Dai, Z., Pasupat, P., Chen, A., Chaganty, A.~T., Fan, Y., Zhao, V.~Y.,
  Lao, N., Lee, H., Juan, D.-C., et~al. (2022).
\newblock Attributed text generation via post-hoc research and revision.
\newblock {\em arXiv preprint arXiv:2210.08726}.

\bibitem[Guinness et~al., 2018]{Guinness2018}
Guinness, D., Cutrell, E., and Morris, M.~R. (2018).
\newblock Caption crawler.
\newblock In {\em Proceedings of the 2018 {CHI} Conference on Human Factors in
  Computing Systems}. {ACM}.

\bibitem[Gurari et~al., 2020]{gurari2020captioning}
Gurari, D., Zhao, Y., Zhang, M., and Bhattacharya, N. (2020).
\newblock Captioning images taken by people who are blind.

\bibitem[Kreiss et~al., 2021]{kreiss2021concadia}
Kreiss, E., Goodman, N.~D., and Potts, C. (2021).
\newblock Concadia: Tackling image accessibility with context.
\newblock {\em arXiv preprint arXiv:2104.08376}.

\bibitem[Lin, 2004]{lin2004rouge}
Lin, C.-Y. (2004).
\newblock Rouge: A package for automatic evaluation of summaries.
\newblock In {\em Text summarization branches out}, pages 74--81.

\bibitem[Lin et~al., 2014]{Lin2014}
Lin, T.-Y., Maire, M., Belongie, S., Hays, J., Perona, P., Ramanan, D.,
  Doll{\'{a}}r, P., and Zitnick, C.~L. (2014).
\newblock Microsoft {COCO}: Common objects in context.
\newblock In {\em Computer Vision {\textendash} {ECCV} 2014}, pages 740--755.
  Springer International Publishing.

\bibitem[Manakul et~al., 2023]{manakul2023selfcheckgpt}
Manakul, P., Liusie, A., and Gales, M.~J. (2023).
\newblock Selfcheckgpt: Zero-resource black-box hallucination detection for
  generative large language models.
\newblock {\em arXiv preprint arXiv:2303.08896}.

\bibitem[Ouyang et~al., 2022]{ouyang2022training}
Ouyang, L., Wu, J., Jiang, X., Almeida, D., Wainwright, C., Mishkin, P., Zhang,
  C., Agarwal, S., Slama, K., Ray, A., et~al. (2022).
\newblock Training language models to follow instructions with human feedback.
\newblock {\em Advances in Neural Information Processing Systems},
  35:27730--27744.

\bibitem[Papineni et~al., 2002]{papineni2002bleu}
Papineni, K., Roukos, S., Ward, T., and Zhu, W.-J. (2002).
\newblock Bleu: a method for automatic evaluation of machine translation.
\newblock In {\em Proceedings of the 40th annual meeting of the Association for
  Computational Linguistics}, pages 311--318.

\bibitem[Radford et~al., 2021]{radford2021learning}
Radford, A., Kim, J.~W., Hallacy, C., Ramesh, A., Goh, G., Agarwal, S., Sastry,
  G., Askell, A., Mishkin, P., Clark, J., et~al. (2021).
\newblock Learning transferable visual models from natural language
  supervision.
\newblock In {\em International conference on machine learning}, pages
  8748--8763. PMLR.

\bibitem[Radford et~al., 2019]{radford2019language}
Radford, A., Wu, J., Child, R., Luan, D., Amodei, D., Sutskever, I., et~al.
  (2019).
\newblock Language models are unsupervised multitask learners.
\newblock {\em OpenAI blog}, 1(8):9.

\bibitem[Ramesh et~al., 2021]{ramesh2021zeroshot}
Ramesh, A., Pavlov, M., Goh, G., Gray, S., Voss, C., Radford, A., Chen, M., and
  Sutskever, I. (2021).
\newblock Zero-shot text-to-image generation.

\bibitem[Robertson, 2004]{Robertson2004}
Robertson, S. (2004).
\newblock Understanding inverse document frequency: on theoretical arguments
  for {IDF}.
\newblock {\em Journal of Documentation}, 60(5):503--520.

\bibitem[Santurkar et~al., 2022]{santurkar2022caption}
Santurkar, S., Dubois, Y., Taori, R., Liang, P., and Hashimoto, T. (2022).
\newblock Is a caption worth a thousand images? a controlled study for
  representation learning.
\newblock {\em arXiv preprint arXiv:2207.07635}.

\bibitem[Schulman et~al., 2017]{schulman2017proximal}
Schulman, J., Wolski, F., Dhariwal, P., Radford, A., and Klimov, O. (2017).
\newblock Proximal policy optimization algorithms.
\newblock {\em arXiv preprint arXiv:1707.06347}.

\bibitem[Thomee et~al., 2016]{Thomee_2016}
Thomee, B., Shamma, D.~A., Friedland, G., Elizalde, B., Ni, K., Poland, D.,
  Borth, D., and Li, L.-J. (2016).
\newblock {YFCC}100m.
\newblock {\em Communications of the {ACM}}, 59(2):64--73.

\bibitem[Vedantam et~al., 2015]{vedantam2015cider}
Vedantam, R., Lawrence~Zitnick, C., and Parikh, D. (2015).
\newblock Cider: Consensus-based image description evaluation.
\newblock In {\em Proceedings of the IEEE conference on computer vision and
  pattern recognition}, pages 4566--4575.

\bibitem[Wang and Komatsuzaki, 2021]{wang2021gptj6b}
Wang, B. and Komatsuzaki, A. (2021).
\newblock Gpt-j-6b: A 6 billion parameter autoregressive language model.

\bibitem[Zhang et~al., 2021]{zhang2021vinvl}
Zhang, P., Li, X., Hu, X., Yang, J., Zhang, L., Wang, L., Choi, Y., and Gao, J.
  (2021).
\newblock Vinvl: Revisiting visual representations in vision-language models.
\newblock In {\em Proceedings of the IEEE/CVF Conference on Computer Vision and
  Pattern Recognition}, pages 5579--5588.

\end{thebibliography}
\bibliographystyle{apalike}

\end{document}